\documentclass[preprint,12pt]{elsarticle}

\usepackage{amssymb}

\journal{Intelligent Data Analysis}

\usepackage[ruled]{algorithm2e}

\usepackage{algcompatible, multirow}
\usepackage{textcomp}

\usepackage{amsthm, amsmath, amssymb}
\usepackage{bm, color}
\usepackage{url}
\usepackage{graphicx, subcaption}
\usepackage{times, epsfig, wrapfig}
\usepackage[flushleft]{threeparttable}
\usepackage{makecell, natbib}
\usepackage{afterpage}

\newtheorem{definition}{Definition}

\newtheorem{assumption}{Assumption}

\newtheorem{problem}{Problem}

\let\OLDthebibliography\thebibliography
\renewcommand\thebibliography[1]{
  \OLDthebibliography{#1}
  \setlength{\parskip}{0pt}
  \setlength{\itemsep}{0pt plus 0.3ex}
}

\begin{document}

\begin{frontmatter}



\title{Discovering Context Specific Causal Relationships}


\author[label1]{Saisai Ma\corref{cor1}}
\ead{saisai.ma@mymail.unisa.edu.au}
\author[label1]{Jiuyong Li}
\author[label1]{Lin Liu}
\author{Thuc Duy Le\fnref{label1}}
\address[label1]{School of Information Technology and Mathematical Sciences, University of South Australia, Mawson Lakes, SA 5095, Australia}
\cortext[cor1]{Corresponding author.}

\begin{abstract}
With the increasing need of personalised decision making, such as personalised medicine and online recommendations, a growing attention has been paid to the discovery of the context and heterogeneity of causal relationships. Most existing methods, however, assume a known cause (e.g. a new drug) and focus on identifying from data the contexts of heterogeneous effects of the cause (e.g. patient groups with different responses to the new drug). There is no approach to efficiently detecting directly from observational data context specific causal relationships, i.e. discovering the causes and their contexts simultaneously. In this paper, by taking the advantages of highly efficient decision tree induction and the well established causal inference framework, we propose the Tree based Context Causal rule discovery (TCC) method, for efficient exploration of context specific causal relationships from data. Experiments with both synthetic and real world data sets show that TCC can effectively discover context specific causal rules from the data.
\end{abstract}


\end{frontmatter}


\newpage
\section{Introduction} \label{sec_intro}
    Causal relationships reveal the causes behind the phenomena and provide insights into the mechanisms of complex systems, therefore finding causal relationships is a central task in many areas. Several causal models, such as causal Bayesian network \cite{pearl_causality:_2000}, structural equation model \cite{w_t_bielby_structural_1977} and potential outcome model \cite{rubin_estimating_1974}, have been proposed to represent and infer causal relationships which are global or context free.

    In reality a variable (e.g. a therapeutic procedure) often has a strong causal effect on an outcome only when the other variables (e.g. genomic profiles) having a specific value. The former variable is called a cause or treatment, while the latter is a context to define a subpopulation. Such causal relationships are called context specific causal relationships in this paper.

    The discovery of context specific causal relationships has important applications in various areas \cite{hyder_national_2010, zsambok_naturalistic_2014, hunink_decision_2014}. For example, for most economical outcomes, it is important to know for different industries (contexts), the most effective polices (causes/treatment) to be implemented. To maximise profit, it is essential to find the customer groups with different shopping profits (contexts) and the profitable products (causes/treatments) for the corresponding groups.

    Context specific causal relationships, however, are hidden and difficult to be discovered since the overall causal effect may be averaged out to be marginal in the whole population. For example, for a treatment, some patients respond positively and some respond negatively, and hence the overall effect among all patients is marginal. A straightforward solution is to assess the treatment effect under all different conditions/contexts, but it is infeasible given the large number of all possible conditions.

    Recently researchers seek to apply data mining and machine learning techniques to the investigation of treatment effect heterogeneity \cite{weisberg_post_2015, athey_recursive_2016}. These techniques are utilised to efficiently find the contexts (subpopulations) across which heterogeneous effects of a treatment can be observed. The work has made it practical to discover the contexts and heterogeneity of causal effects.

    However, from the data mining perspective, these techniques bear a major limitation. They assume a known cause (i.e. treatment) variable and focus on finding the proper contexts where the cause has heterogeneous causal effects on the outcome. Therefore, it is not suitable for the exploration for context specific causal relationships in data, where the causes are unknown.
    With the assumption of a known cause relaxed, a big challenge arises for finding context specific causal relationships directly from data, that is, how to distinguish potential causal/treatment variables from context variables. 

    Our goal is to design a data mining method to discover context specific causal relationships without knowing or assuming a cause, that is, to find both contexts and the causal relationships under the contexts simultaneously. Our approach to this challenge, TCC (Tree based Context Causal rule discovery) adapts decision tree induction, like the work in~\cite{athey_recursive_2016}, but in a very different way. In \cite{athey_recursive_2016}, a causality based criterion is used to build a causal tree for finding the subpopulations across which a treatment has heterogeneous effects.
    Instead we directly make use of the highly efficient and mature decision tree algorithm~\cite{Quinlan1993} to find candidate causes and context variables with respect to a given target. Then within the much reduced search space, we employ the potential outcome model \cite{rubin_estimating_1974} to assess the candidates to identify causes and their contexts.

    We use decision tree as a base for the following two reasons. Firstly, a rational assumption is that contexts and causes are all highly related to the target, so it is reasonable to use decision tree to select the candidates. Meanwhile, each decision rule encodes context specific relationships between predictor variables and the target, which are likely the indicators of context specific causal relationships. Secondly, a decision tree is efficient for both large sized and high dimensional data, and hence basing TCC on decision tree induction will be practical for various applications. In contrast, it is multiple orders of magnitude slower to build a causal tree compared to a normal decision tree, as the causality based criterion is performed in each split of the tree construction to examine each variable for choosing the optimal branching variable~\cite{athey_recursive_2016, li_causal_2016}.

    We further extend decision rules to context specific causal rules for actionable decision making. For example, along a path in a decision tree, a decision rule like $(X_1=1,X_2=1) \rightarrow (Y=1)$ shows the co-occurrence of $(X_1=1,X_2=1)$ and $Y=1$, which is sufficient for classification. However, such a rule is insufficient for actions, since it is important to know which variable leads to the change of $Y$ in actionable decision making. For example, in biomedical experimental design, the decision rule can be interpreted as ``$X_1 \rightarrow Y | X_2=1$" or ``$X_2 \rightarrow Y | X_1=1$", which means totally different manipulation operations. The former refers to manipulating $X_1$ under the context $X_2=1$, while the latter is to manipulate $X_2$ when $X_1=1$. Therefore, causality based examination is in demand to identify the causes and their contexts for evidence based decision making.

    We take this work truly as a journey of causal knowledge discovery from large data sets, therefore our method design has been focused on practical approach and the TCC algorithm has been aimed at quickly finding meaningful causal signals and their contexts in a large data set. The experimental results have shown that TCC performs consistently when it is applied to synthetic or real world data sets, and its high efficiency is also proved by the experiments.

     One significance of our work is that we demonstrate that a supervised learning method can be easily adapted for causal discovery with high efficiency and high quality.

    In the rest of this paper, the problem statement is presented in Section \ref{sec_problem}, and then a practical definition of context specific causal rules is defined under the potential outcome model. The proposed method is discussed in Section \ref{sec_disc}. Section \ref{sec_simulate} demonstrates the performance of the proposed method. Section \ref{sec_related} reviews related work. Finally, we conclude the paper in Section \ref{sec_conclusion}.

{ 
\section{Related Work} \label{sec_related}

    Many attentions have been paid to causal discovery on observational data.
    Various causal models have been developed for causal relationship discovery \cite{rubin_bias_1980, pearl_causality:_2000, ma_mining_2016, li_causal_2016}.
    The potential outcome model \cite{rubin_estimating_1974} has been widely used for the estimation of causal relationships. Matching methods \cite{stuart_matching_2010} are developed to remove confounding when estimating the average causal effect of the treatment on the outcome. Rosenbaum and Rubin \cite{rosenbaum_central_1983} proposed the propensity score matching for average causal effect estimation, where a logistic regression is used to estimate the propensity score. 


    A growing literature focuses on modelling and finding context specific relationships. A stream of research is to derive Context Specific Independence (CSI) based on a known Bayesian network \cite{boutilier_context-specific_1996, koller_probabilistic_2009}. Researchers intended to speed up the Bayesian network inference algorithms by introducing the concept of CSI. Instead of aiming at fast inference with Bayesian networks, some others focused on extending a Bayesian network by adding special notations, such as labelled graphical models \cite{corander_labelled_2003}, gates \cite{minka_gates_2009} and stratified Gaussian graphical models \cite{nyman_stratified_2014}, such that the extended Bayesian network can explicitly present the context specific causal relationships. These methods normally assume that global dependency relationships between variables are known in advance.

    Another main stream of research that is related to context specific causal discovery is subgroup analysis. Subgroup analysis is commonly used to evaluate the treatment effects in a specific subpopulation defined by some context variables. Su et al. \cite{su_subgroup_2009} adapted the idea of recursive partitioning to construct an interaction tree for the causal effect estimation. Dudik et al. \cite{dudik_doubly_2011} developed an approach to get the optimal policy via the technique of doubly robust estimation. Supervised machine learning approaches have been applied to estimate heterogeneous causal effects \cite{weisberg_post_2015, athey_recursive_2016}.


    However, these methods are designed to validate hypothesised causal effects of subgroups and the hypotheses have been provided based on the domain knowledge at the commencement of a study. The subjective hypotheses may result in that previously unobserved patterns and relationships would never be tested. What we expect is not only to validate the hypothesised causal relationships, but also to find unobserved causal relationships previously. Thus computational methods are required to discover causal relationships from observational data automatically.

    Causal decision tree method \cite{li_causal_2016} was developed to explore both general and context specific causal relationships. Specifically, the causal relationship between the root node and the outcome is context free, while non-root nodes are causes of the outcome under the context of their parent nodes. Although such type of trees have widely practical applications, it has a limitation that the contexts of causes have to be already causes (or context specific causes) of the outcome.
}

\section{Problem Statement and Definitions} \label{sec_problem}
    In this section, we firstly state the research problem of this work, then we define context specific causal rules, and discuss how to identify a context specific causal rule from data.

    \subsection{Research problem}
    The objective of the work is to find context specific causal relationships in data. Specifically, we aim to find context specific causal rules as stated below.
    \begin{problem}
        Given a data set $\bm{D}$ with a set of predictor variables $\bm{V}$ and the target variable $Y$, find all the potential treatment variables $X_p \in \bm{V}$ and the corresponding context variables $\bm{X}_c$ ($\bm{X}_c \subset \bm{V} \backslash X_p$), such that $X_p \rightarrow Y$ is a causal rule when $\bm{X}_c=\bm{x}_c$.
    \end{problem}


    A rule $X \rightarrow Y$ is causal, if the treatment variable $X$ has a significant causal effect on the target variable $Y$, that is, varying $X$ will result in a significant change of $Y$. In other words, a causal rule $X \rightarrow Y$ satisfies two major conditions: ($i$) the variable $X$ precedes the target $Y$, and ($ii$) if $X$ had not happened, $Y$ would be different.

    The first condition specifies a temporal relationship between variables $X$ and $Y$, which normally can be identified with domain knowledge. In our study, we always assume that all treatment variables precede the outcome temporally. 
    The second condition is at the conceptual level and it indicates that outcome $Y$ would be different when the same individual received a treatment and did not receive it. The difference between the two outcomes under treatment and no treatment is typically called the treatment/causal effect \cite{rubin_bias_1980, pearl_causality_2000}. 


    Similarly, we have the following criteria of identifying a context specific causal rule. A rule $X_p \rightarrow Y | \bm{X}_c=\bm{x}_c$ is a causal rule in the context $\bm{X}_c=\bm{x}_c$, if ($i$) the treatment variable $X_p$ and context variables $\bm{X}_c$ are disjunctive, i.e. $X_p \cap \bm{X}_c = \emptyset$, and ($ii$) within the context $\bm{X}_c=\bm{x}_c$, $X_p$ has a significant causal effect on $Y$. In a special case, $\bm{X}_c$ can be an empty set, and thus the context specific causal rule becomes a general causal rule (i.e. context free).

    In the next section, we will formally present a practical definition of causal rules and context specific causal rules, and the estimation of causal effects.

    \subsection{Causal rule definition} \label{sec_instant}
    The potential outcome model \cite{rubin_estimating_1974, morgan_counterfactuals_2014} is widely used in the estimation of causal effects in social science, health and medical research. In this model, an individual $i$ in a population has two potential outcomes with respect to a treatment (we only consider binary treatment in this paper): when taking the treatment ($X_p=1$), the potential outcome is $Y_1(i)$; and when not taking the treatment ($X_p=0$), the potential outcome is $Y_0(i)$.

    However, for an individual $i$, we can only observe one of the two potential outcomes, either $Y_1(i)$ or $Y_0(i)$. The unobserved outcomes, namely the counterfactual outcomes, need to be estimated by using the observed outcomes, such that we can compare the difference of outcomes when receiving treatment or control. 
    
    The individual level causal effect is expressed as $Y_1(i) - Y_0(i)$. The causal effects of individuals in a population are normally aggregated to get the Average Causal Effect (ACE) as defined below:
    \begin{align}\label{eq_ACE}
    ACE(X_p \rightarrow Y) = E[Y_1] - E[Y_0]
    \end{align}
    where $E[.]$ stands for the expectation operator in probability theory. Note that $i$ is omitted when we focus on the population level potential outcomes.

    With the definition of ACE, the practical definitions of causal rules and context specific causal rules are formally presented in the following.

    \begin{definition} [Causal rules] \label{def_causal_instntiation}
        Given a data set $\bm{D}$ with a set of predictor variables $\bm{V}$ and the target variable $Y$, a rule $X_p \rightarrow Y$ ($X_p \in \bm{V}$) is a causal rule, if $ACE(X_p \rightarrow Y) \ge \eta$ in $\bm{D}$, where $\eta$ is the minimal causal effect threshold.
    \end{definition}

    \begin{definition} [Context specific causal rules] \label{def_contxC}
        Given a data set $\bm{D}$ with a set of predictor variables $\bm{V}$ and the target variable $Y$, a rule $X_p \rightarrow Y | \bm{X}_c = \bm{x}_c$ ($X_p \in \bm{V}, \bm{X}_c \subset \bm{V}$ and $X_p \cap \bm{X}_c = \emptyset$) is a causal rule in the context $\bm{X}_c = \bm{x}_c$, if $ACE(X_p \rightarrow Y | \bm{X}_c = \bm{x}_c) \ge \eta$, where $\eta$ is the minimal causal effect threshold.
    \end{definition}

    The threshold $\eta$ can be determined based on domain knowledge.
    
    {Note that in this paper we assume that the differences of individuals could be captured by the covariates, i.e. the set of variables used for stratification. This assumption implies that there are no hidden confounding variables to bias the causal effect estimation.}

    \subsection{Causal effect estimation}
    The major issue for ACE estimation is to unbiasedly estimate the counterfactual outcomes, e.g. what the effect would be if a person had not taken a treatment (actually the person did take the treatment). If we have two groups of individuals, one group taking a treatment and another not, and the two groups of individuals have the same characteristics apart from being treated or not, we can straightforwardly estimate the counterfactual outcomes based on the observed outcomes. In this process, the indistinguishability of two groups apart from treated or not is essential. 
    
    Randomised treatment assignment is a way to achieve indistinguishability. However, with observational data, such random assignments of treatments are often not guaranteed. In this case, stratification of the data set is a way of trying to achieve the indistinguishability. In each stratified sub data set, the records of all covariates take the same values in the treatment ($X=1$) and control ($X=0$) groups, respectively. Thus under the stable unit treatment value assumptions \cite{rubin_bias_1980}, the individuals of the two groups in a stratum are indistinguishable, except the state of the treatment. Then in each stratum, we can unbiasedly estimate the counterfactual outcomes and obtain ACE.

    Now we present the details of the procedure of causal effect estimation with observational data.

    \subsubsection{Variables used for stratification}
    The first step of causal effect estimation is to determine the set of covariate variables (denoted by $\bm{C}$ in the paper) to be used for stratifying data. In a non-experimental study, a key assumption for the variable selection is the unconfoundedness \cite{rosenbaum_central_1983}:

    \begin{assumption}
        The treatment assignment $X$ is independent of the potential outcomes ($Y_0$, $Y_1$) given the covariates $\bm{C}$, i.e. $X \perp (Y_0, Y_1) | \bm{C}.$
    \end{assumption}

	\afterpage{%
    \begin{figure}[!p]
        \center
        \includegraphics[width=0.5\textwidth]{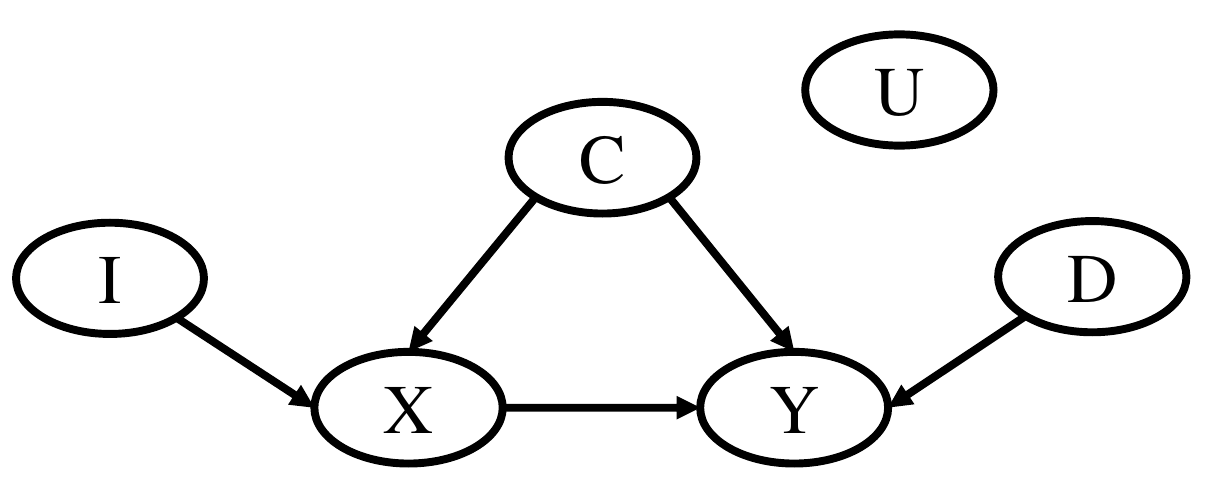}
        \caption{A causal diagram}
        \label{fig_causalFrame}
    \end{figure}
    \clearpage
	}

    The causal diagram in Figure \ref{fig_causalFrame} is used to help with the following discussions of covariate selection. In this figure, the nodes represent variables and the edges denote the causal links between the nodes\footnote{{ Note that a causal diagram is different with an influence diagram, where an arrow denotes an influence and it does not necessarily imply a causal relation.}}. For example, the connection $X \rightarrow Y$ means $X$ is a cause of the target $Y$. Apart from treatment $X$ and the target $Y$, the other variables are categorised into four different types: ($i$) Indirect causes (e.g. $I$), which indirectly cause a change of $Y$; ($ii$) Confounders (e.g. $C$), which are the common causes of $X$ and $Y$; ($iii$) Direct causes (e.g. $D$), which are direct causes of $Y$, apart from $X$; and ($iv$) Irrelevant variables (e.g. $U$), which are totally independent with both $X$ and $Y$.

    From this causal structure, we can see that confounders $C$ are the ones that may influence the causal effect estimation of the treatment $X$ on $Y$. Thus to satisfy the assumption of unconfoundedness, variables known to have causal effects on both treatment assignment and the outcome, i.e. the confounders $C$ shown in Figure \ref{fig_causalFrame}, are required to be included in stratification \cite{stuart_matching_2010}. Unfortunately, the causal graph is typically unknown. In this paper, all variables that are associated with both the treatment variable $X$ and the target $Y$ are included into covariates $\bm{C}$ for stratification, as a variable can never be a cause of another if they are independent. Covariates $\bm{C}$ are normally a superset of confounders $C$ in Figure \ref{fig_causalFrame}. With the propensity score method used in this paper (details in Section 2.3.3), it has been shown that there is less cost in terms of increased bias to include variables that actually do not impact on the causal effect estimation of $X$ on $Y$, compared to the case excluding potentially important confounders \cite{stuart_matching_2010}.

    \subsubsection{Distance measures and stratification}
    The second step is to choose a distance measure for stratification. Perfect stratification (i.e. all samples in a stratum have exactly same values), is ideal to eliminate the bias, but it does not work when the number of covariates is large since the statistical power is lost quickly with the increase of the number of covariates. To improve the statistical power, approximate stratifications are developed to match individuals with similar covariate distributions (not exact ones).
    
    { Various distance measure, e.g. Minkowski distance and Mahalanobis distance, can be used for the stratification, but most of them does not perform well when there are many covariates under study \cite{gu_comparison_1993}.} Propensity score \cite{rosenbaum_central_1983, stuart_matching_2010} is another commonly used distance measure, which summarises covariates $\bm{C}$ into {one scalar:} the probability of the individual receiving the treatment conditioning on $\bm{C}$:
    \begin{align}\label{eq_propensity}
        e(\bm{C}) = Prob(X=1 | \bm{C}=\bm{c}).
    \end{align}


    Subclassification on propensity score~\cite{stuart_matching_2010} is used here to do stratification, i.e. grouping individuals with similar propensity scores to a stratum, such that individuals are indistinguishable (in terms of receiving the treatment or not) within one stratum.

    \subsubsection{Causal effect estimation}
    After the data set has been stratified based on propensity scores, we can estimate the causal effect within each stratum and the aggregate the causal effects over the strata to obtain the overall causal effect. In each stratum $\bm{C} = \bm{c}_k$, a contingency table, shown in Table \ref{tab_contingecy}, is generated for the estimation of average causal effect. $a$, $b$, $c$ and $d$ are the counts of variable $X_p$ and the outcome $Y$ with different values, and $n_k=a+b+c+d$ is the number of samples/individuals in the sub data set with the context $\bm{C} = \bm{c}_k$.

	\afterpage{%
    \begin{table}
    	\footnotesize
      \centering
      \caption{An example of notations for a contingency table}\label{tab_contingecy}
    \begin{tabular}{cccc}
        \hline
        $X_p$              & $Y=1$     & $Y=0$     &  Total \\
        \hline
        1              & $a$  & $b$  & $a+b$ \\
        0   & $c$  & $d$  & $c+d$ \\
        \hline
        Total               & $a+c$  & $b+d$  & $n_k$ \\
        \hline
    \end{tabular}
    \end{table}
    \clearpage
	}

    Referring to the definition of ACE, the causal effect is the difference of the outcomes in two groups. Thus in the stratum $\bm{c}_k$, the average causal effect is expressed as
    \begin{align}\label{eq_ACEstratum}
        ACE(X_p \rightarrow Y | \bm{c}_k) = \frac{a}{a+b} - \frac{c}{c+d}.
    \end{align}

    The ACE in a population is determined by aggregating the ACEs in all strata
    \begin{align}\label{eq_ACEaggre}
        ACE(X_p \rightarrow Y) = \sum_{k} w_k ACE(X_p \rightarrow Y | \bm{c}_k).
    \end{align}
    where $w_k$ is the weight of the stratum $\bm{C}=\bm{c}_k$. In this paper, $w_k$ is set as the ratio of the sample size of $\bm{c}_k$ to the size of data $\bm{D}$.

\section{Context Specific Causal Rule Discovery} \label{sec_disc}
    In this section, we firstly present the proposed algorithm, TCC, for mining context specific causal rules with a single decision tree, then we introduce a variant of TCC to explore context specific causal rules with multiple trees.

    \subsection{TCC with a single decision tree} \label{subsec_causalSingle}
        As shown in Algorithm~\ref{alg}, TCC contains two major parts: decision rule selection (lines 1 to 7) and causal rule discovery with a pruning strategy (lines 8 to 25).

        TCC firstly picks up a proper search base for finding causal rules by learning a decision tree from the data. C4.5~\cite{Quinlan1993} is employed to build a decision tree from data. We restrict the minimum number of instances per leaf, such that there are enough samples for the ACE estimation. Each path in the decision tree is a decision rule $\bm{R}$ expressed as $(\bm{X}=\bm{x}) \rightarrow (Y=y)$ or $\bm{x} \rightarrow y$, where $\bm{X}$ and $Y$ are the predictor variables and the target on a path of the decision tree, and $\bm{x}$ and $y$ are the corresponding values respectively.

\afterpage{%
        \begin{algorithm}[t]
        \caption{Tree based Context Causal rule discovery (TCC) algorithm} \label{alg}
            \textbf{Input:} A data set $\bm{D}$ for predictor variable set $\bm{V}$ and the target $Y$, the minimal confidence threshold $\theta$, and the minimal causal effect threshold $\eta$. \\
            \textbf{Output:} $\bm{C_Y}$, the set of causal rules to the target $Y$.
            \begin{algorithmic}[1]
                \STATEx // Building single or multiple decision trees
                \STATE $\bm{T} = \emph{decisionTree}(\bm{V},Y)$
                \STATE $\bm{C_Y} = \emptyset$
                \FOR{each decision rule $\bm{R}$ of decision trees $\bm{T}$}
                    \STATE extract the predictor variables $\bm{X}$ and $Y$ and their
                    \STATEx \hspace{\algorithmicindent} corresponding values $\bm{x}$ and $y$ from $\bm{R}$
                    \IF{$\emph{confTest}(\bm{X},Y) \le \theta$}
                        \STATE \textbf{continue}
                    \ENDIF
                    \STATEx \hspace{\algorithmicindent} // Global causal test
                    \FOR{each variable $X_p \in \bm{X}$}
                       \STATE $\emph{ACEValue} = \emph{causalTest}(\bm{D}, X_p, \emptyset, \emptyset, Y)$
                       \IF{$\emph{ACEValue} > \eta$}
                            \STATE $\bm{C_Y} = \bm{C_Y} \cup \{X_p \rightarrow Y\}$
                       \ENDIF
                    \ENDFOR
                    \STATEx \hspace{\algorithmicindent} // Context specific causal test
                    \FOR{each variable $X'_p \in \bm{X}$}
                       \STATE $\bm{X'} = \bm{X} \backslash \{X'_p\}$
                       \FOR{each variable set $\bm{X}_c \subset \bm{X'}$, $\bm{X}_c=\bm{x}_c$}
                          \IF{$redundantTest(X'_p,\bm{X}_c, \bm{x}_c,\bm{C_Y})$}
                             \STATE \textbf{continue}
                          \ENDIF
                          \STATE $\emph{ACEValue} = \emph{causalTest}(\bm{D}, X'_p, \bm{X}_c, \bm{x}_c, Y)$
                          \IF{$\emph{ACEValue} > \eta$}
                          \STATE $\bm{C_Y} = \bm{C_Y} \cup \{X'_p \rightarrow Y | \bm{X}_c=\bm{x}_c\}$
                          \ENDIF
                       \ENDFOR
                    \ENDFOR
                \ENDFOR
                \STATE Output $\bm{C_Y}$
            \end{algorithmic}
        \end{algorithm}
    \clearpage
	}

        To guarantee the statistical significance, we also use the Fisher's exact test to prune branches of a decision tree \cite{liu_robust_2010}. With the notation in Table \ref{tab_contingecy}, $X_p$ and $Y$ here refers to a branching variable and the outcome, and the $p$-value is given by:
        $$p([a,b;c,d])=\sum_{i=0}^{min(b,c)}{\frac{(a+b)!(c+d)!(a+c)!(b+d)!}{n_k!(a+i)!(b-i)!(c-i)!(d+i)!}}$$

        A low $p$-value means that the null hypothesis (i.e. $X_p$ and $Y$ are independent) is rejected. We only keep branches that are statistically significant (with low $p$-values).

        Given all decision rules of a decision tree, the predictor variables in each decision rule are considered as the search base of both potential causes and contexts, as a decision rule encodes context specific relationships. Then a confidence test (line 5 in Algroithm~\ref{alg}) is conducted to remove the decision rules if it has low confidence, since causal signal in a low confidence rule is weak. Here the confidence of $\bm{x} \rightarrow y$ is defined as the proportion of individuals containing $\bm{x}$ which also contains $y$. Only a decision rule with high confidence, i.e. exceeding the specified minimal confidence threshold, will be inserted into the candidate set for causal rule discovery.

        For a high confidence decision rule $\bm{x} \rightarrow y$, global causal tests are performed to detect if $X_p \rightarrow Y$ ($X_p \in \bm{X}$) is a global causal rule. Lines 6 to 9 show this process, where Formula \ref{eq_ACE} is employed to estimate the ACE.

        Then we move to context specific causal rule discovery. The discovery of context specific causal rules from a decision rule includes two nested loops (lines 10 to 17 in Algorithm \ref{alg}). In the outer loop, we traverse the predictor variables $\bm{X}$ in $\bm{R}$ as the candidate treatment variable $X'_p$, while the inner loop enumerates the subsets of $\bm{X} \backslash \{X'_p\}$ finding the contexts. With each subset $\bm{X}_c$ of $\bm{X} \backslash \{X'_p\}$, the subset of data is extracted from the original data with $\bm{X}_c=\bm{x}_c$, where $\bm{x}_c$ is the value of $\bm{X}_c$ as indicated in the decision rule $\bm{R}$.

        A bottle-neck for context specific causal rule discovery is the enumeration of different contexts in the variable set of the antecedent of a decision rule. Thus a pruning strategy is developed to address the efficiency problem. Function \emph{RedundantTest}() (line 13) is invoked to test if the rule $X_p \rightarrow Y | \bm{X}_c=\bm{x}_c$ is redundant. Only if the rule is not redundant, then causal test (line 14) is performed to estimate the causal effect of $X'_p$ on $Y$ under the context $\bm{X}_c=\bm{x}_c$.

        As we know, if a causal relationship holds in a population, then it should hold in each of the subpopulations. In other words, if $X_p \rightarrow Y | \bm{X}_c=\bm{x}_c$ is a context specific causal rule, $X_p \rightarrow Y | \{\bm{X}_c=\bm{x}_c,\bm{X_a}=\bm{x_a}\}$ is also a context specific causal rule, where $\bm{X_a}$ is an additional condition defining a specific subpopulation. The more specific rules (i.e. with more conditions than the general one) are implied by the general causal rule. For example, if Children's Panadol is effective for relieving child under 12 from fever and pain (i.e. $Panadol \rightarrow recovery | age<12$), then we can conclude that it is also effective for boys under 12 (i.e. $Panadol \rightarrow recovery | \{age<12, sex=male\}$). We call such more specific rules as redundant rules.

        We are not interested in redundant rules as the causal relationships (if any), since the causal relationships are already implied by their more general context specific causal rules. Thus we exclude redundant rules in the algorithm to reduce the search space. Once we find a context-specific causal rule (including a global causal rule where the context variable set $X_c=\emptyset$), we stop searching for its more specific context specific rules.

    \subsection{TCC with multiple decision trees}
        The performance of TCC could be sensitive to the results of decision tree construction. A decision tree normally makes use of a small subset of variables in the decision rules, so a key limitation of using decision tree for our purpose is that it may not cover all possible causal factors and the context variables, and thus we may miss some potential causal relationships. In this section, we present a variant of TCC with an ensemble classifier, Diversified Multiple Trees (DMT) \cite{li_building_2016}, to address the false negative issue.

        DMT uses C4.5 \cite{Quinlan1993} to sequentially build $m$ decision trees, where attributes used in a tree are not to be used in the construction of the next tree. Thus the output decision trees are disjunct. Then decision rules extracted from the output DMTs are used as the search space of the TCC algorithm.

        To avoid confusion, we call the TCC algorithm with DMT as ``TCC$_m$", where $m$ is the number of decision trees built, and ``TCC" without a subscript denotes the TCC algorithm with a single tree as introduced in \ref{subsec_causalSingle}.

        As DMT is capable of detecting more attributes highly correlated with the target, potentially TCC$_m$ has less false negatives and is expected to achieve higher accuracy.

    \subsection{Complexity analysis} \label{subsec_TimeC}
        The time complexity of the proposed method TCC comes from three main parts: tree construction, general causal rule extraction, and context specific causal rule extraction. Here we focus on analysing the performance of context specific causal extraction, since the complexity of two other parts is significantly lower than the complexity of this part. We denote the height of a tree as $h$, the number of variables as $m$, and the number of samples as $n$.

        The number of paths of the tree is $2^{h'-1}$ where $h' < h$ considering that not all paths have the same length of $h$. For each path, we enumerate the contexts of all variables along the path and we have $2^{h''-1}$ possible contexts where $h'' \ll h$ because of the effect of pruning. The total number of context specific tests is in the order of $O(2^{\beta h})$ where $ 0.5 < \beta < 2$.  In each test, finding covariates is at the cost of $O(m^2)$. For computing the propensity score, the complexity ranges from $O(n\log(n))$ (regression tree) to $O(n^{\alpha})$ (logistic regression) where $2 \le \alpha \le 3$~\cite{komarek_logistic_2004}. The overall complexity is between $O(2^{\beta h} (n\log(n) + m^2))$ and $O(2^{\beta h} (n^{\alpha} + m^2))$. Consider that $h$ is normally a small integer and $m \ll n$. The complexity can be approximately in the order of $O(l*n\log(n))$ where $l$ is in the range of hundreds to thousands.

\section{Experiments} \label{sec_simulate}
    In this section, we firstly introduce the process of synthetic data generation. Then we present the experiments on TCC and TCC$_m$ with the synthetic data sets, and compare the performance of TCC and TCC$_m$ with the Causal Tree (CT) method \cite{athey_recursive_2016}. CT is designed to examine the heterogeneity of causal effects across subsets of the population, while assuming known cause. Specifically, it applies regression tree with a modified MSE (Mean Squared Error) criterion to partition the population into multiple subgroups.
    
    Then we apply TCC and TCC$_m$ to a clinical data set, the METABRIC data set, to capture meaningful context specific causal rules. 

    \subsection{Synthetic data} \label{subsec_synGenerate}
        In order to evaluate the proposed method, we generate several synthetic data sets containing context specific causal relationships.

        Each of the synthetic data sets is generated with four main steps: ($i$) create randomly two Causal Bayesian Networks (CBNs) with the same number of variables by using the TETRAD software tool (http://www.phil.cmu.edu/tetrad/), where the Direct Acyclic Graphs (DAGs) and Conditional Probability Tables (CPTs) are both created randomly by the software; ($ii$) generate two data sets from the two causal Bayesian networks based on their respective conditional probability tables via the built-in Bayes Instantiated Model; ($iii$) add one more column to each of two data sets, as the context variable $X_c$, such that $X_c \equiv 0$ in the first data set and $X_c \equiv 1$ in the other one; and ($iv$) concatenate these two new data sets by columns to obtain the final data set.

        We use the above procedure to generate five synthetic data sets with 10, 20, 30, 40 and 50 variables (Syn-10, Syn-20, Syn-30, Syn-40, and Syn-50) respectively. Each data set also has 10K samples. Then we use precision ($P$), recall ($R$) and $F_1$-measure ($F_1$) as the metrics to evaluate the performance of TCC, TCC$_m$ and CT in term of the accuracy. Different from TCC, CT focuses on a fixed treatment variable to estimate the differences in causal effects of the treatment across different subpopulations, where the treatment variable is a hypothesised cause of the target variable. For comparing with our proposed method, we conduct multiple independent runs of the CT algorithm, with each predictor variable set as a treatment. Here we only discuss the results within the context $X_c = 0$ and $X_c = 1$, since all we know about the ground truth are the causes in the context $X_c$.

	\afterpage{%
        \begin{table}
        \footnotesize
        \centering
        \caption{The accuracy of TCC, TCC$_3$, TCC$_5$ and CT on synthetic data sets.}
        \label{tab_compare.20}
            \begin{tabular}{c|ccc|ccc|ccc|ccc}
            \hline
                & \multicolumn{3}{c|}{TCC} & \multicolumn{3}{c|}{TCC$_3$} & \multicolumn{3}{c|}{TCC$_5$} & \multicolumn{3}{c}{CT} \\
                & $P$ & $R$ & $F_1$ & $P$ & $R$ & $F_1$ & $P$ & $R$ & $F_1$ & $P$ & $R$ & $F_1$ \\
            \hline
            Syn-10    & 0.83  & 0.83  & 0.83 & 0.83  & 0.83  & 0.83 & 0.83  & 0.83  & 0.83 & 0.75 & 1.00 & \textbf{0.86} \\
            Syn-20    & 1.00  & 0.83  & \textbf{0.91} & 1.00  & 0.83  & \textbf{0.91} & 1.00  & 0.83  & \textbf{0.91} & 0.50 & 0.75 & 0.60 \\
            Syn-30    & 1.00  & 0.83  & \textbf{0.91} & 1.00  & 0.83  & \textbf{0.91} & 1.00  & 0.83  & \textbf{0.91} & 0.38 & 1.00 & 0.55 \\
            Syn-40    & 1.00  & 1.00  & \textbf{1.00} & 1.00  & 1.00  & \textbf{1.00} & 0.86  & 1.00  & 0.92 & 0.29 & 0.67 & 0.40 \\
            Syn-50    & 1.00  & 0.83  & \textbf{0.91} & 0.83  & 0.83  & {0.83} & 0.83  & 0.83  & {0.83} & - & - & - \\
            \hline
            \end{tabular}
        \end{table}
    \clearpage
	}

        The results discovered by TCC, TCC$_3$, TCC$_5$ and CT are shown in the Table \ref{tab_compare.20}. We can see that CT has achieved high performance on the small data set, Syn-10, while its performance drops sharply as the number of variables increases. CT becomes infeasible on a larger data set, Syn-50, as the efficiency of CT is sensitive to both dimension and size of a data set. In contrast, TCC, TCC$_3$ and TCC$_5$ consistently achieve high performance and the average of \emph{$F_1$ score} is larger than 0.80. Meanwhile, these three methods obtain very similar results. It is because on these five data sets, single decision tree has included almost all potential causes and corresponding contexts, and thus TCC with multiple trees could not make more improvement.

	\afterpage{%
        \begin{table}
    	\footnotesize
        \centering
        \caption{Performance of TCC and TCC$_m$ on Syn-100.}
        \label{tab_DMT}
            \begin{tabular}{c|cccc}
            \hline
                & TCC & TCC$_3$ & TCC$_5$ & TCC$_7$ \\
            \hline
            $P$    & 0.80  & 0.82  & 0.90 & 0.90 \\
            $R$    & 0.80  & 0.90  & 0.90 & 0.90 \\
            $F_1$    & 0.80  & 0.86  & 0.90 & 0.90 \\
            \hline
            \end{tabular}
        \end{table}
    \clearpage
	}

        To further examine the performance of TCC and TCC$_m$, we run TCC and TCC$_m$ ($m=3$, 5 and 7) on a larger data set, Syn-100. Table \ref{tab_DMT} shows the results with Syn-100. The accuracy is significantly improved when DMT is employed, and more decision trees (larger the values of $m$) bring bigger improvement. This is due to the fact that multiple trees may be able to include more potential causes and context variables than a single decision tree, which in turn improves the chance of detecting context specific causes. When all possible causal factors and context variables are included in the output trees, the accuracy improvement will stop, which explains why TCC$_7$ and TCC$_5$ get the same $F_1$ score.

    \subsection{Scalability}
        To evaluate the efficiency of TCC and TCC$_m$, we run experiments on five synthetic data sets with 50, 100, 150, 200 and  250 variables, respectively. The experiments are performed on a computer with a 3.4 GHz Quad-core CPU and 16 GB of memory.

	\afterpage{%
        \begin{figure}
            \centering
            \begin{subfigure}[c]{0.48\textwidth}
                \centering
	           \includegraphics[scale=0.36]{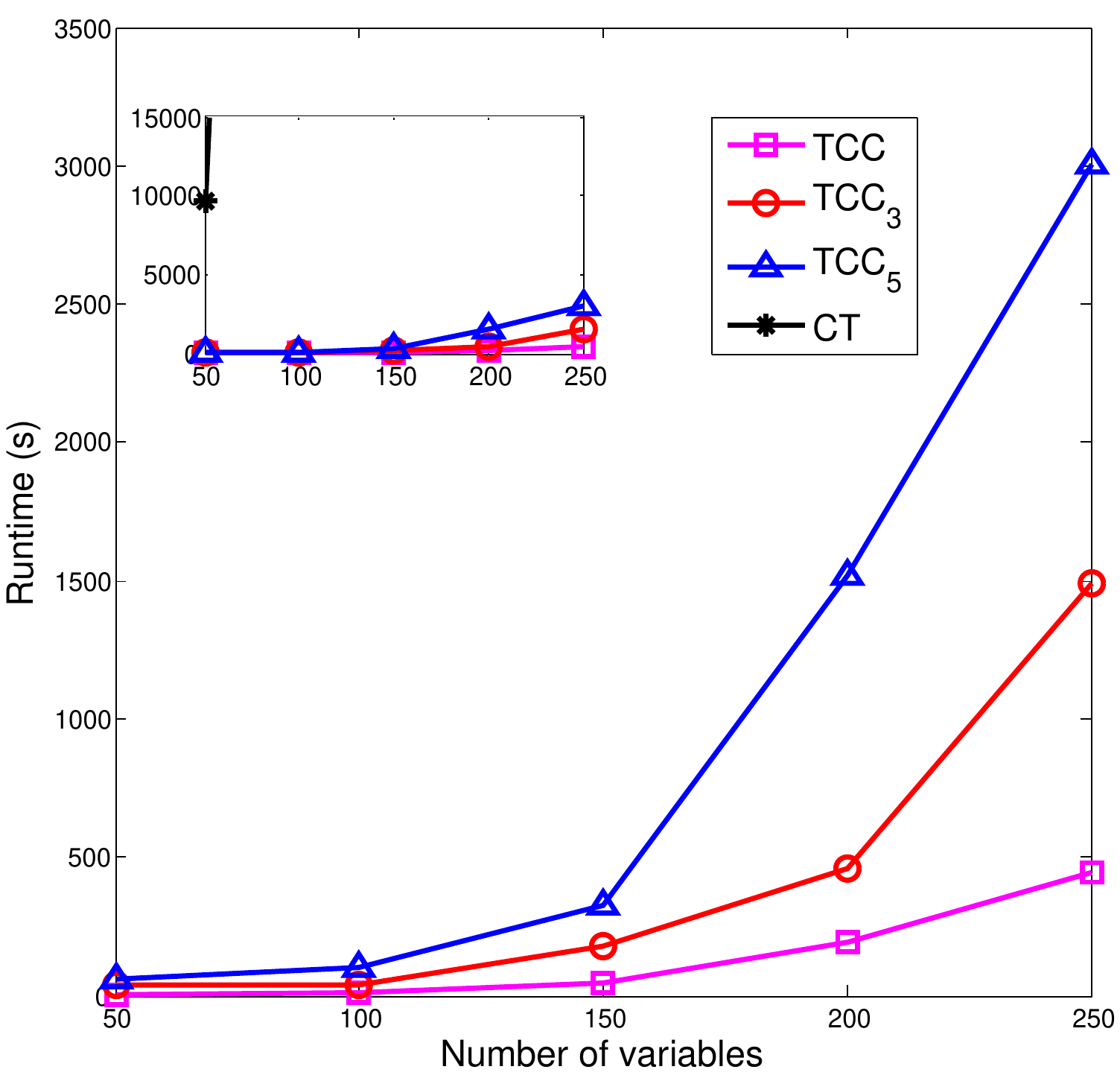}
                \caption{\# context variables = 1, i.e. $|\bm{X}_c|=1$}
            \end{subfigure}
            \begin{subfigure}[c]{0.48\textwidth}
                \centering
	           \includegraphics[scale=0.36]{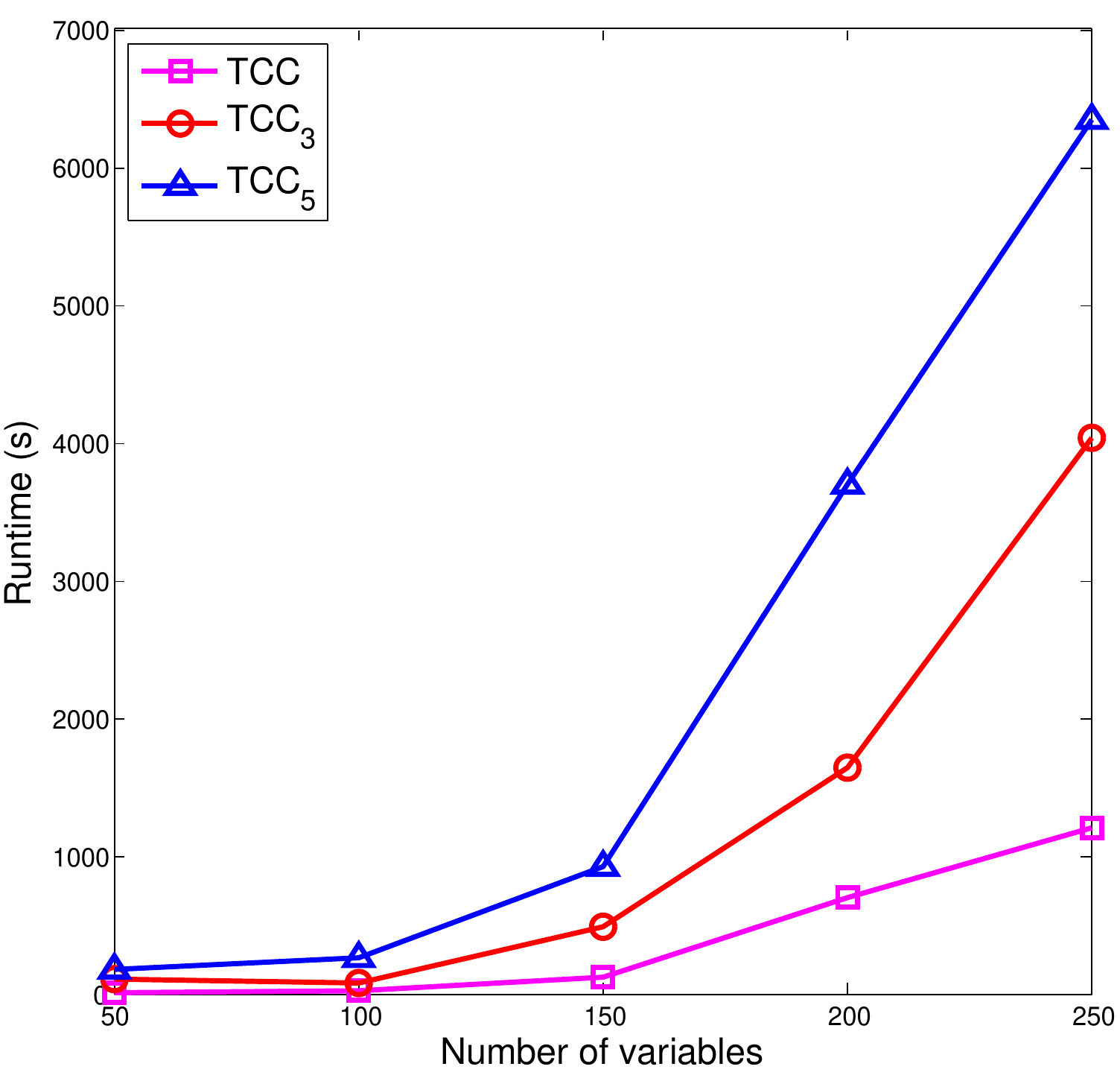}
                \caption{\# context variables = 2, i.e. $|\bm{X}_c|=2$}
            \end{subfigure}
            \caption{Scalability evaluation with TCC and TCC$_m$.}
            \label{fig_scale}
        \end{figure}
    \clearpage
	}

        We run TCC, \emph{TCC$_3$} and \emph{TCC$_5$} with $|\bm{X}_c|=1$ and $|\bm{X}_c|=2$ respectively, where $|\bm{X}_c|$ means the cardinality of the context variable set, and compare the running time with CT. Note that only the results returned within 5 hours are shown in Figure \ref{fig_scale}. From Figure \ref{fig_scale}(a) we see that TCC is the most efficient algorithm, while the run time of TCC$_m$ (here $m=3,5$) is roughly $m$ times of TCC. CT is much slower than our proposed methods and it is not feasible with large data sets. The results show that CT is not competent in exploring for context specific causal relationships in large data sets. Figure \ref{fig_scale}(b) shows the efficiency of TCC, TCC$_3$ and TCC$_5$, when the size of context variables is set as 2. All three methods are also efficient and scalable. Specifically, in this setting, TCC and TCC$_m$ require slightly more than double of the time used when $|\bm{X}_c|=1$.

    \subsection{METABRIC data set} \label{sec_application}
    	We also apply the proposed method to real world data, the METABRIC data set \citep{soulakis_visualizing_2015}. The data set contains clinical traits and outcomes for 1981 primary breast cancer patients collected from participants of the Molecular Taxonomy of Breast Cancer International Consortium (METABRIC) trial. We run the proposed method on the data set with two different outcomes, 10 years Overall Survival (OS) and 10 years Disease Free Survival (DFS), respectively. The 10 years OS status indicates whether a patient died from breast cancer within 10 years or is alive 10 years after initial consultation. The 10 years DFS status indicates whether a patient survives more than 10 years or not without any signs or symptoms of breast cancer after the cancer ends due to the primary treatment. The METABRIC data set contains 570 and 762 patients who have positive and negative 10 years OS status, respectively, and 820 and 516 patients with positive and negative 10 years DFS status, respectively. { Note that to avoid unexpected noise or/and incorrectness involved, instead of imputing missing data \cite{azur_multiple_2011}, we directly removed the records with missing values from the METABRIC data set.}
    	
    	Firstly, the decision tree method is employed to detect association based relationships with respect to breast cancer. Then each decision rule is considered as the search base of both potential causes and contexts. In the set of experiments with this real world data set, available domain knowledge can be utilised to further reduce the search space, to improve the computational efficiency, that is, a potential cause could only be one of three different treatment therapies: chemotherapy, hormone therapy and radiotherapy. Note that these three therapies could also be the contexts to define specific subpopulations, who received multiple treatments.
    	
	\afterpage{%
    	\begin{table}
    	\footnotesize
    	\centering
    	\caption{Top context specific causal rules discovered by TCC from METABRIC with 10 years OS as the outcome.}
    	\label{tab_OS-1}
    	\begin{tabular}{c|ccc}
    	\hline
    	Treatment & Context & 10 years OS & Reference \\
    	\hline
    	\multirow{4}{*}{Chemotherapy} & Breast surgery = breast conserving & Yes & \cite{sanford_impact_2016} \\ \cline{2-4}
    	& \makecell{Cellularity = high \& \\ Nottingham prognostic index  $\le$ 5} & No &  \\ \cline{2-4}
    	& Cellularity = low & No &  \\
    	\hline
    	\end{tabular}
    	\end{table}
    \clearpage
	}
    	
    	We firstly run TCC on the METABRIC data set with 10 years OS as the outcome, and extract top context specific causal rules based on causal effects, shown in Table \ref{tab_OS-1}. The causal rules discovered by TCC are supported (or partially supported) by domain knowledge and literature. The first rule in the table indicates that chemotherapy is very effective within the subpopulation, where patients received a lumpectomy to remove a part of the breast tissue, instead of the entire breast \cite{sanford_impact_2016}. Some interesting context specific causes are also detected. The other two significant causal rules are showing that chemotherapy is not an effective treatment for the subpopulation that has low tumor cellularity mass and the subpopulation with high tumor cellularity mass and low nottingham prognostic index.
    
	\afterpage{%
    	\begin{table}
    	\footnotesize
    	\centering
    	\caption{Top context specific causal rules discovered by TCC from METABRIC with 10 years DFS as the outcome.}
    	\label{tab_DFS-1}
    	\begin{tabular}{c|ccc}
    	\hline
    	Treatment & Context & 10 years DFS & Reference \\
    	\hline
    	\multirow{5}{*}{Chemotherapy} & \makecell{IntClust $>$ 2 \& \\ Claudin subtype = claudin-low} & Yes & \cite{ali_genome-driven_2014, prat_phenotypic_2010} \\ \cline{2-4}
    	& Pam50 subtype = normal & Yes & \cite{prat_response_2015} \\ \cline{2-4}
    	& Age $>$ 60 & No \\ \cline{2-4}
    	& \makecell{IntClust $>$ 2 \& \\ Claudin subtype = not classified} & No & \cite{ali_genome-driven_2014} \\
    	\hline
    	\multirow{4}{*}{\makecell{Hormone \\ therapy}} & \makecell{Chemotherapy = no \& \\ Three gene = ER+/HER2- \\ high proliferation} & Yes & \cite{rastelli_factors_2008} \\ \cline{2-4}
    	& \makecell{Inferred menopausal state = post \& \\ Radiotherapy = no} & Yes \\ \cline{2-4}
    	& Three gene = ER-/HER2- & No & \cite{rastelli_factors_2008} \\
    	\hline
    	\multirow{7}{*}{Radiotherapy} & Hormone therapy = yes & Yes \\ \cline{2-4}
    	& \makecell{Chemotherapy = no \& \\ Age $\le$ 60} & Yes & \cite{mao_revisiting_2017} \\ \cline{2-4}
    	& \makecell{Chemotherapy = no \& \\ Pam50 subtype = HER2} & No & \cite{mao_revisiting_2017} \\ \cline{2-4}
    	& \makecell{Chemotherapy = no \& \\ Pam50 subtype = not classified} & No \\
    	\hline
    	\end{tabular}
    	\end{table}
    \clearpage
	}
    	
    	We then set 10 years DFS as the outcome, and run the proposed method on the data set. Some interesting and reasonable results are found (Table \ref{tab_DFS-1}). For example, the third rule in the table shows that chemotherapy has a negative impact on older people (Age $>$ 60). TCC also confirms the effectiveness of chemotherapy when IntClust $>$ 2 and Claudin subtype = claudin-low \cite{prat_phenotypic_2010}, where the IntClust approach \cite{dawson_new_2013} classifies the breast cancer into ten subtypes based on gene expression and it is good for survival when IntClust is larger than 2 \cite{ali_genome-driven_2014}. The results also show that hormone therapy is poor to cure the patients with lower Estrogen Receptors (ER-) \cite{rastelli_factors_2008}. Radiotherapy cannot bring survival benefit to the patients with HER2 Pam50 subtype \cite{mao_revisiting_2017}.
    	
	\afterpage{%
    	\begin{table}
    	\footnotesize
    	\centering
    	\caption{Top context specific causal rules discovered by TCC$_3$ from METABRIC with 10 years OS as the outcome.}
    	\label{tab_OS-3}
    	\begin{tabular}{c|ccc}
    	\hline
    	Treatment & Context & 10 years OS & Reference \\
    	\hline
    	Chemotherapy & Claudin subtype = claudin-low & No & \cite{prat_phenotypic_2010} \\
    	\hline
    	\multirow{3}{*}{\makecell{Hormone \\ therapy}} & \makecell{Tumor size $\le$ 57 \& \\ Chemotherapy = no} & Yes \\ \cline{2-4}
    	& HER2 SNP6 = loss & No \\ \cline{2-4}
    	& Claudin subtype = Luminal-A & No \\
    	\hline
    	\end{tabular}
    	\end{table}
    \clearpage
	}
    	
    	We also apply TCC with three trees (TCC$_3$) on the data set with two different outcome variables, 10 years OS and 10 years DFS, respectively. For the data set with 10 years DFS, TCC and TCC$_3$ have captured similar causal rules with strong causal effects, so here we only show the results of TCC$_3$ on the data with 10 years OS as the outcome. Comparing to the TCC results, more causal rules are discovered and some examples are shown in Table \ref{tab_OS-3}. Patients with claudin-low tumors have poor overall survival outcomes, even if they received chemotherapy \cite{prat_phenotypic_2010}. Hormone therapy is beneficial to the patients with small tumor size.

\section{Conclusion} \label{sec_conclusion}
    In this paper, a novel method, Tree based Context Causal rule discovery (TCC) has been proposed to explore context specific causal relationships from observational data. Finding causes and contexts simultaneously is important but challenging. Decision tree is utilised to make the complex problem manageable. We have designed TCC based on a well-known causal framework, the potential outcome model, to assess context specific causal relationships. A variant, TCC$_m$ (i.e. TCC with multiple decision trees) is also introduced to help improve the performance of TCC. 

    The experiments results show that TCC can achieve high performance with the synthetic data sets and find insights from real world data sets. TCC also outperforms an existing causal tree method, in terms of the exploration of short and meaningful context specific causal relationships and easy operation without specifying a candidate cause. TCC provides a scalable and automated way to address the increasing need of uncovering context specific causal relationships for personalised decision making.

{\section{Acknowledgement}
This work has been supported by Australian Research Council (ARC) Discovery Project Grant DP170101306 and the National and Medical Research Council (NHMRC) Grant 1123042.}



\end{document}